\documentclass[11pt,a4paper]{article}
\usepackage{coling2020}
\usepackage{times}
\usepackage{latexsym}

\usepackage[hang,flushmargin]{footmisc}  
\usepackage{amsfonts,amsmath,amssymb}
\usepackage[T1]{fontenc}  
\usepackage{bold-extra}   
\usepackage{microtype}    
\usepackage{subcaption}
\usepackage{graphicx}
\usepackage{multirow}
\usepackage{enumitem}
\usepackage{url}

\colingfinalcopy 

\title{XD at SemEval-2020 Task 12: Ensemble Approach to Offensive Language Identification in Social Media Using Transformer Encoders}

\author{Xiangjue Dong \\
  Computer Science \\
  Emory University \\
  Atlanta, GA, USA \\
  {\tt xiangjue.dong@emory.edu} \\\And
  Jinho D. Choi \\
  Computer Science \\
  Emory University \\
  Atlanta, GA, USA \\
  {\tt jinho.choi@emory.edu} \\}

\date{}

\begin{document}
\maketitle
\begin{abstract}
This paper presents six document classification models using the latest transformer encoders and a high-performing ensemble model for a task of offensive language identification in social media. 
For the individual models, deep transformer layers are applied to perform multi-head attentions. 
For the ensemble model, the utterance representations taken from those individual models are concatenated and fed into a linear decoder to make the final decisions. 
Our ensemble model outperforms the individual models and shows up to 8.6\% improvement over the individual models on the development set. On the test set, it achieves macro-F1 of 90.9\% and becomes one of the high performing systems among 85 participants in the sub-task A of this shared task.
Our analysis shows that although the ensemble model significantly improves the accuracy on the development set, the improvement is not as evident on the test set.

\end{abstract}
\section{Introduction}
\label{sec:introduction}
%
%
\blfootnote{
    %
    %
    %
    %
    %
    %
    This work is licensed under a Creative Commons 
    Attribution 4.0 International License.\\
    License details: \url{http://creativecommons.org/licenses/by/4.0/}.
}

With the development of IT, social media has become more and more popular for people to express their views and exchange ideas publicly. However, some people may take advantage of the anonymity in social media platform to express their comments rudely, and attack other people verbally with offensive language. To keep a healthy online environment for the adolescences~\cite{chen-etal:2012} and to filter offensive messages for the users~\cite{razavi-etal:2010}, it is necessary and significant for technology companies to develop an efficient and effective computational methods to identify offensive language automatically. 



Transformer-based contextualized embedding approaches such as \texttt{BERT} \cite{devlin_2019}, \texttt{XLNet} \cite{yang_2019a}, \texttt{RoBERTa} \cite{liu_2019}, \texttt{ALBERT} \cite{lan_2019} or \texttt{ELECTRA} \cite{clark_2020} have re-established the state-of-the-art for many natural language classification tasks especially the \textsc{GLUE} Dataset \cite{wang_2018}. Their pre-trained models were pre-trained on different large datasets, for example, \texttt{BERT} was pre-trained on the \textsc{BookCorpus} \cite{zhu_2015} and English Wikipedia, and \texttt{RoBERTa} was pre-trained on \textsc{CC-News} \cite{nagel_2016}, \textsc{OpenWebText} \cite{gokaslan_2019}, and \textsc{Stories} \cite{trinh_2018} which enable their models to learn different language features.

This paper presents six transformer-based offensive language identification models that learn different features from the target utterance. To combine the distinctive learned language features, we introduce an ensemble strategy which concatenates the representations of the individual models and feed them into the linear decoder to make binary classification (Section~\ref{ssec:models}). It largely improves the performance over the baseline on our dev set (Section~\ref{ssec:results}).

\section{Related Work}
\label{sec:related-work}

Offensive language in Twitter~\cite{wiegand-etal:2018}, Facebook~\cite{kumar-etal:2018}, and Wikipedia~\cite{georgakopoulos-etal:2018} has been widely studied. In addition, different aspects of offensive language have been studied, like the type and target of offensive posts~\cite{zampieri-etal:2019}, cyberbullying~\cite{dinakar-etal:2011,huang-etal:2014}, aggression~\cite{kumar-etal:2018}, toxic comments~\cite{georgakopoulos-etal:2018} and hate speech~\cite{badjatiya-etal:2017,davidson-etal:2017,malmasi-zampieri:2017,malmasi-zampieri:2018}. 

Many deep learning approaches have been used to address the task. The Convolutional Neural Networks (CNNs), Long Short-Term Memory Networks (LSTMs) and FastText were applied on the hate speech detection task~\cite{badjatiya-etal:2017}. Gamback and Sikdar~\shortcite{gamback-sikdar:2017} used four Convolutional Neural Network (CNN) models with random word vectors, word2vec word vectors, character n-gram, and concatenation of word2vec word embeddings and character n-grams as feature embeddings separately to categorize each tweet into four classes: racism, sexism, both (racism and sexism) and non-hate-speech.

\section{Data Description}
\label{sec:task-description}
The datasets we use are Offensive Language Identification Dataset (\texttt{OLID})~\cite{zampieri-etal:2019} and Semi-Supervised Offensive Language Identification Dataset (\texttt{SOLID}) \cite{rosenthal-etal:2020}. 
Given a tweet, the task is to predict whether the content involves offensive language.
Table~\ref{tab:examples} shows the examples of offensive and non-offensive tweets in these two datasets.

\begin{table}[htbp!]
\centering\small

\begin{subtable}{\columnwidth}
\centering\small
\begin{tabular}{c||c|c}
\bf Id & \bf Tweet & \bf Label \\
\hline\hline
09 & @USER Buy more icecream!!! & \texttt{NOT} \\
\hline
71 & @USER That's because you are an old man. & \texttt{OFF}
\end{tabular}
\caption{Examples from \texttt{OLID}.}
\label{stab:old-example}
\end{subtable}
\vspace{1ex}

\begin{subtable}{\columnwidth}
\centering\small
\resizebox{\columnwidth}{!}{
\begin{tabular}{c||c|c|c}
\bf Id & \bf Tweet & \bf \texttt{AVG\_CONF} & \bf \texttt{CONF\_STD} \\ 
\hline\hline
167 & @USER Pre-ordered your book, received in July, started last night and cannot put it down!
& 0.215 & 0.188 \\
\hline
524 & a combination of innocence and corruption & 0.691 & 0.142 \\
\end{tabular}}
\caption{Examples from \texttt{SOLID}.}
\label{stab:new-example}
\end{subtable}

\caption{Examples in \texttt{OLID} and \texttt{SOLID}. \texttt{NOT}: not offensive, \texttt{OFF}: offensive, \texttt{AVG\_CONF}: average of the confidences to be offensive, \texttt{CONF\_STD}: confidences' standard deviation}
\label{tab:examples}
\end{table}

\noindent \texttt{OLID} is a collection of 14,100 English tweets annotated as \texttt{OFF} or \texttt{NOT}. 
It is divided into a training set of 13,240 tweets and test set of 860 tweets~\cite{zampieri-etal:2019}. 
\texttt{SOLID} is a collection of about 9 million English tweets labeled in a semi-supervised manner~\cite{rosenthal-etal:2020}. 
The data are annotated with \texttt{AVG\_CONF} and \texttt{CONF\_STD} predicted by several supervised models~\cite{zampieri-etal:2020}.
The test set provided by organizers this year has 3887 tweets. 
Table~\ref{tab:statistics} shows the statistics of \texttt{OLID} and \texttt{SOLID}.
\begin{table}[htbp!]
\centering\small 
\begin{tabular}{c||c|c}
 & \bf \texttt{OLID} & \bf \texttt{SOLID} \\
 \hline\hline
 \texttt{TRN} & 13240 & 9089140 \\
 \texttt{TST} & 860 & 3887 \\
\end{tabular}
\caption{Statistics of \texttt{OLID} and \texttt{SOLID}. \texttt{TRN}: training set, \texttt{TST}: test set.}
\label{tab:statistics}
\end{table}

\section{Experiments}
\label{sec:experiments}

\subsection{Data Split}
\label{ssec:data-split}

For our experiments, a combination of \texttt{OLID} and \texttt{SOLID} (Section~\ref{sec:task-description}) is used. 
We find that about 1.0\% of \texttt{SOLID} are duplicates, which have been removed before data splitting. 
For the dataset used for fine-tuning classification model, we set threshold of \texttt{AVG\_CONF} (Section~\ref{sec:task-description}) to be 0.5 in \texttt{SOLID}, which means the data with \texttt{AVG\_CONF} above 0.5 is labelled as \texttt{OFF}. 
90\% of the \texttt{TRN} of \texttt{OLID} is combined with the whole \texttt{SOLID} as the new training set \texttt{TRN} for default transformer-based models fine-tuning (\texttt{FT}).
The remaining 10\% of the \texttt{TRN} and the \texttt{TST} of \texttt{OLID} is used as the development set \texttt{DEV} of \texttt{FT}. 
All the existed datasets are combined together as the training set \texttt{TRN} for model pre-training (\texttt{PT}). 
After pre-training, 99.5\% of the \texttt{SOLID} is randomly selected as the training set \texttt{TRN} and 0.5\% of the \texttt{SOLID} is randomly selected to create the development set \texttt{DEV} for fine-tuning our pre-trained models into classification models and regression models (\texttt{PT-C} and \texttt{PT-R}). In \texttt{PT-C}, the data with \texttt{AVG\_CONF} above 0.5 is labelled as \texttt{OFF} and in \texttt{PT-R}, original value of \texttt{AVG\_CONF} is used. Furthermore, 90\% of \texttt{TRN} in \texttt{OLID} is randomly selected as the new training set \texttt{TRN}, and 10\% of \texttt{TRN} in \texttt{OLID} is combined with the \texttt{TST} of the \texttt{OLID} and become the development set \texttt{DEV} for classification models and regression models' further fine-tuning (\texttt{PT-C-C} and \texttt{PT-R-C}). 
The ensemble model is fine-tuned on the same dataset as \texttt{PT-C-C}.
Table~\ref{tab:train-dev} shows the detailed statistics of the data split in our experiments. 

\begin{table}[htbp!]
\centering\small 
\begin{tabular}{c||c|c|c|c|c|c|c}
 & \texttt{FT} & \texttt{PT} & \texttt{PT-R} & \texttt{PT-C} & \texttt{PT-R-C} & \texttt{PT-C-C} & \texttt{E} \\
\hline\hline
\texttt{TRN} & 8,963,663  & 9,107,127 & 8,951,747 & 8,951,747 & 11,916 & 11,916 & 11,916 \\
\texttt{DEV} & 2,184 & - & 44,983 & 44,983 & 2,184 & 2,184 & 2,184
\end{tabular}
\caption{Statistics of the data split used for our experiments. \texttt{TRN}: training set, \texttt{DEV}: development set, \texttt{FT}: dataset used for default model fine-tuning, \texttt{PT}: dataset used for default model pre-training, \texttt{PT-R}: dataset used for fine-tuning our pre-trained models into regression models, \texttt{PT-C}: dataset used for fine-tuning our pre-trained models into classification models, \texttt{PT-R-C}: dataset used for fine-tuning regression model into classification models, \texttt{PT-C-C}: dataset used for further fine-tuning classification models, \texttt{E}: dataset used for fine-tuning ensemble models.}
\label{tab:train-dev}
\vspace{-2ex}
\end{table}

\subsection{Models}
\label{ssec:models}

In general, default transformer-based models are fine-tuned as baseline models. The sequence of embeddings of input generated from the transformer encoder is fed into linear decoder to gain the output vector that makes the binary classification.
Then we pre-train these default models and choose the models with lowest perplexity. 
Next, we fine-tune the pre-trained models into regression models and classification models based on corresponding dataset, respectively. 
Furthermore, the regression models and classification models are fine-tuned again into classification models. 
In the end, sentence presentation of individual models are concatenated and fed into linear decoder to generate the output vector that makes the binary decision of whether or not this tweet is offensive.

\begin{figure*}[htbp!]
\centering
\includegraphics[scale=0.3]{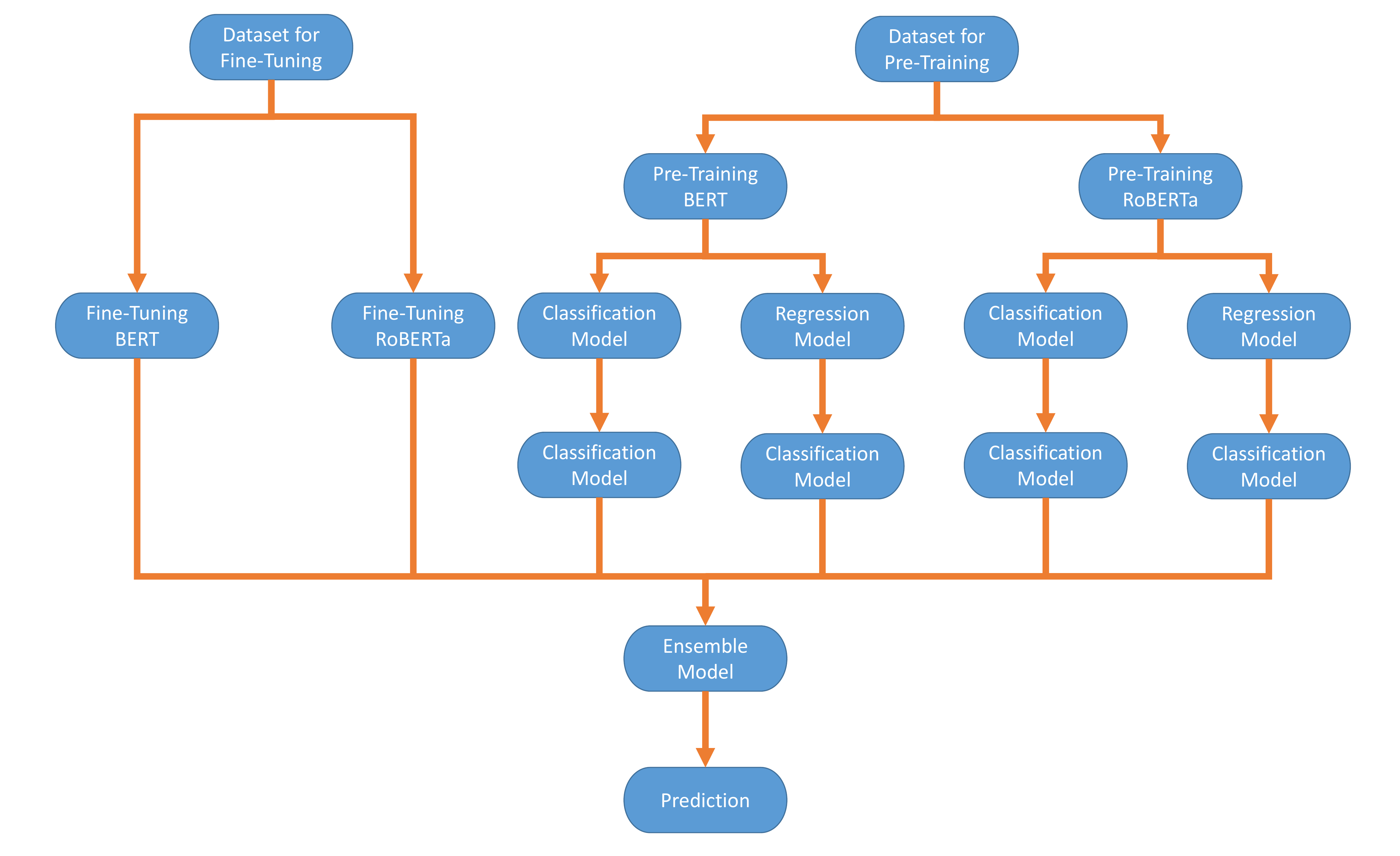}
\caption{Overview of the individual models and the ensemble model}
\label{fig:model-overview}
\end{figure*}

\noindent In our experiments, two types of transformer-based models are used as the default models, BERT-Base model \cite{devlin-etal-2019-bert} and RoBERTa-Base model \cite{liu-etal-2020-roberta}. 
For the default model fine-tuning part, the default BERT-Base and RoBERTa-Base are fine-tuned on \texttt{FT} (Section~\ref{ssec:data-split}) as baseline models. 
For the pre-training part, the BERT-Base and RoBERTa-Base are pre-trained on \texttt{PT} (Section~\ref{ssec:data-split}). 
Then, the two pre-trained models which have the lowest perplexity are fine-tuned into regression models and classification models separately on \texttt{PT-R} and \texttt{PT-C}. 
Next, the fine-tuned pre-trained models are further fine-tuned into classification models on \texttt{PT-R-C} and \texttt{PT-C-C}. 
Finally, sentence presentation of six individual models are concatenated to form the ensemble model which is fine-tuned on \texttt{E}. 
Figure~\ref{fig:model-overview} shows the overview of the six individual models and the ensemble model.

\subsection{Experimental Setup}
\label{ssec:experimental-setup}

According to our experiments, the data preprocessing doesn't contribute significantly to the final prediction results on such huge dataset.
Thus, we skip the data preprocessing. 
According to the analysis of sentence length in the dataset, we set \textit{max\_length} of the models to be $128$. 
After an extensive hyper-parameter search, we set \textit{learning\_rate} to be $2e-5$, \textit{seed\_value} to be $42$, and \textit{epochs} to be $10$ for our six individual models and ensemble model. After that, we also experiment more on the ensemble model and find that the best result is gained by changing \textit{learning\_rate} to $1e-5$ and \textit{dropout} to $0.5$.

\subsection{Results}
\label{ssec:results} 

Table~\ref{tab:results} shows the results achieved by our individual models and ensemble model. 
The selected pre-trained BERT-base model and pre-trained RoBERTa-base model have the lowest perplexities, which are 21.3 and 47.5.
Our fine-tuned pre-trained classification-classificaion BERT and RoBERTa models outperform their counterpart baseline by about 1.7\% and 1.1\%, respectively.
In addition, our fine-tuned pre-trained regression-classification BERT and RoBERTa models show 2.1\% and 1.8\% improvements over their baselines.
The ensemble model with \textit{learning\_rate} of $1e-5$ and \textit{dropout} of 0.5 (\texttt{E\_2}) achieves significantly improvement on development set. 
It outperforms the BERT baseline and RoBERTa baseline by 8.5\% and 8.6\%, respectively. 
As a result, we use this ensemble model as our final model and submit the prediction results to the shared task's CodaLab page.\footnote{\url{https://competitions.codalab.org/competitions/23285}}
We achieve a macro-F1 score of 90.901\% on the test set and rank 36th among 85 participants in sub-task A. 
After the release of the gold labels, we also calculate our other models' performance on test set (Table~\ref{tab:results}) and make detailed comparison and analysis among them (Section~\ref{ssec:ablation-analysis}).

\begin{table}[htbp!]
\centering\small
\begin{tabular}{c||c|c|c|c|c|c}
\bf Model & \bf \texttt{ACC\_DEV} & \bf \texttt{ACC\_TST} & \bf \texttt{P\_TST} & \bf \texttt{R\_TST} & \bf \texttt{F1\_TST} & \bf Epochs \\
\hline
\hline
\texttt{B-FT} & 83.784 & \bf 92.153 & 88.990 & 94.510 & \bf 90.933 & 6 \\
\texttt{R-FT} & 83.692 & 92.102 & 88.933 & 94.503 & 90.882 & 10 \\
\texttt{B-PT-C-C} & 85.204 & 90.610 & 88.402 & 88.115 & 88.256 & 1 \\
\texttt{B-PT-R-C} & 85.845 & 92.102 & 88.933 & 94.532 & 90.885 & 2 \\
\texttt{R-PT-C-C} & 84.654 & 92.102 & 88.933 & 94.532 & 90.885 & 1 \\
\texttt{R-PT-R-C} & 85.158 & 88.552 & 85.129 & 87.886 & 86.299 & 2 \\
\texttt{E} & 88.548 & \bf 92.153 & 88.990 & 94.510 & \bf 90.933 & 2 \\
\texttt{E\_1} & 90.701 & \bf 92.153 & 88.992 & 94.396 & 90.917 & 1 \\
\texttt{\bf E\_2} & \bf 90.884 & 92.128 & 88.962 & 94.464 & 90.901 & 2 \\
\end{tabular}
\caption{\label{tbl:results} Results of individual models and ensemble model on dev set and test set.
\texttt{B-FT}: fine-tuned default BERT-base,
\texttt{R-FT}: fine-tuned default RoBERTa-base,
\texttt{B-PT-C-C}: fine-tuned our pre-trained BERT-base classification-classification model,
\texttt{R-PT-C-C}: fine-tuned our pre-trained RoBERTa-base classification-classification model,
\texttt{B-PT-R-C}: fine-tuned our pre-trained BERT-base regression-classification model,
\texttt{R-PT-R-C}: fine-tuned our pre-trained RoBERTa-base regression-classification model,
\texttt{E}: ensemble model with default \textit{learning\_rate} of $2e-5$, 
\texttt{E\_1}: ensemble model with lower \textit{learning\_rate} of $1e-5$, 
\texttt{E\_2}: submitted ensemble model with higher \textit{dropout} of $0.5$.}
\label{tab:results}
\end{table}

\subsection{Analysis}
\label{sec:analysis}

\subsubsection{Ablation Analysis}
\label{ssec:ablation-analysis}

When we fine-tuned our pre-trained models, \texttt{B-PT-C}, \texttt{B-PT-R}, \texttt{R-PT-C}, and \texttt{R-PT-R} on only 10\% of the \texttt{PT-R} and \texttt{PT-C} (Section~\ref{ssec:results}) separately, the accuracy of models, \texttt{B-PT-C-C}, \texttt{B-PT-R-C}, \texttt{R-PT-C-C}, and \texttt{R-PT-R-C} we get is 82.822\%, 83.326\%, 83.280\%, and 83.646\%, which is lower than the results using total data (Table~\ref{tab:results}).
It indicates that deep learning models which are trained on larger dataset perform better. 
For the ensemble model, when we decrease the \textit{learning\_rate} from $2e-5$ (\texttt{E}) to $1e-5$ (\texttt{E\_LL}), the performance improves from 88.548\% to 90.701\%, which shows that the ensemble model is sensitive to the change in learning rates. 
By changing the default \textit{dropout} from 0.1 (\texttt{E\_LL}) to 0.5 (\texttt{E\_HD}), the model performance increase to 90.884\%, which indicates the influence of the dropout rate. 
After comparing the predicted labels from our unsubmitted models with the released gold labels (Table~\ref{tab:results}), we can see the model which achieves the highest accuracy on the development set doesn't perform best on the test set. which may be caused by overfitting.
Pure fine-tuned BERT-base model (\texttt{B\_FT}) achieves the same accuracy as other two ensemble models.
In addition, higher accuracy can't guarantee the higher f1-score due to the data imbalance.

\subsubsection{Error Analysis}
\label{ssec:error-analysis}
\begin{figure*}[htbp!]
\centering
\includegraphics[scale = 0.35]{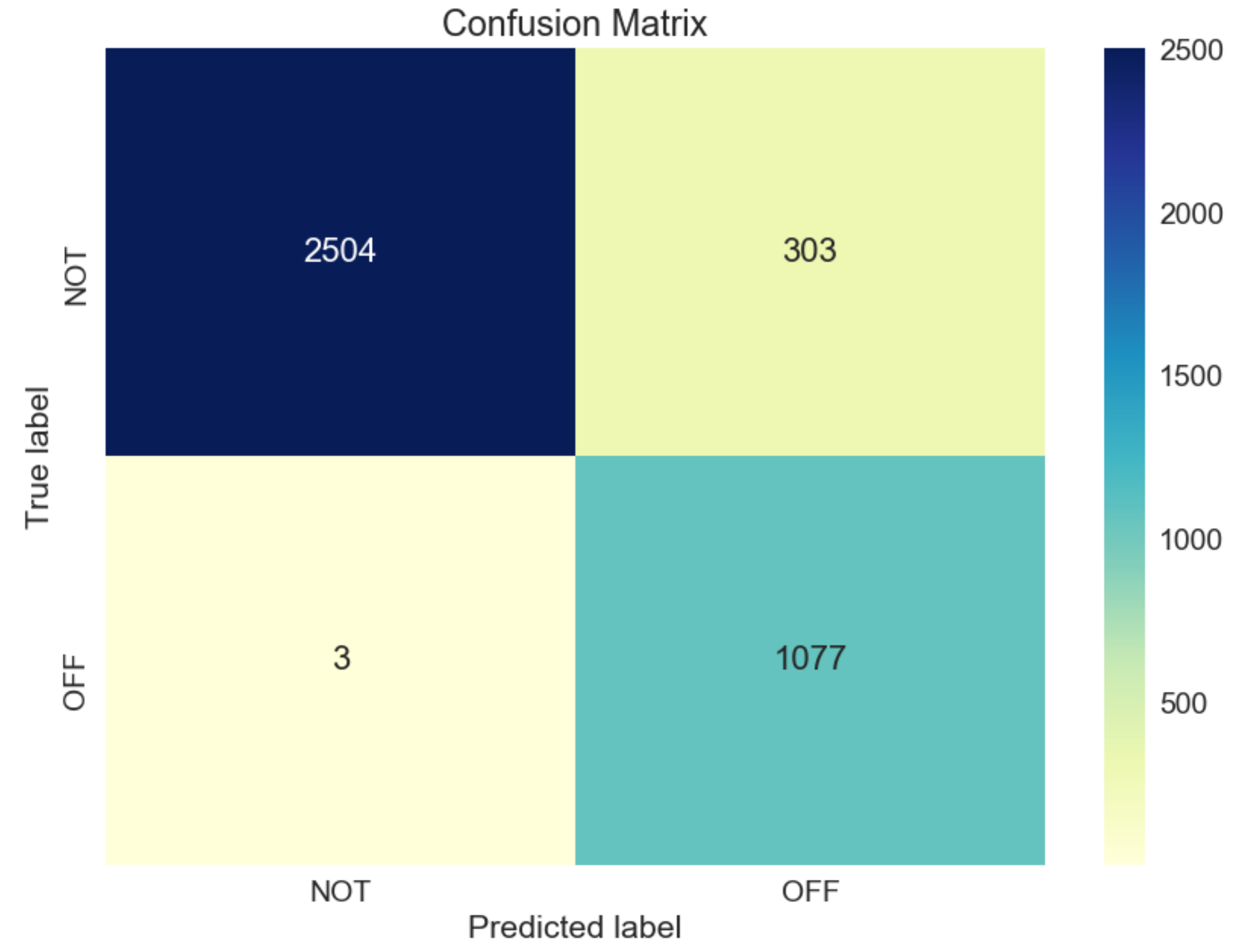}
\caption{Confusion matrix of the ensemble model}
\label{fig:confusion-matrix}
\end{figure*}

The confusion matrix in Figure~\ref{fig:confusion-matrix} further displays the error pattern of our classifier on test set. 
As we can see, there are only three instances labeled with \texttt{OFF} are misclassified to \texttt{NOT} while more data labeled with \texttt{NOT} are classified to \texttt{OFF}.
Table~\ref{tab:misclassified-examples} shows these three misclassified offensive examples and other misclassified not offensive tweets.

\noindent One explanation of the results may be that the imbalance of the dataset leads to the classifier’s preference for the majority class. 
It is possible that our classifier may not capture some of the subtle nuances in meaning and contexts, and our system still needs some improvement for these subtle details.

\begin{table}[htbp!]
\centering\small

\begin{subtable}{\columnwidth}
\centering\small
\resizebox{\columnwidth}{!}{
\begin{tabular}{c||c|c|c}
\bf Id & \bf Tweet & \bf \texttt{PL} & \bf \texttt{TL} \\
\hline\hline
304 & Can someone please jump her ass. & \texttt{NOT} & \texttt{OFF} \\
\hline
2333 & @USER I don’t  So far as you can recognize your Dad aa trash there’s no need to be talking to you. & \texttt{NOT} & \texttt{OFF} \\
\hline
2825 & @USER Wings over and it’s not even a question (sweet chili \&amp; Jamaican jerk hanger). & \texttt{NOT} & \texttt{OFF}
\end{tabular}}
\caption{Misclassified offensive examples.}
\label{stab:misoff-example}
\end{subtable}
\vspace{1ex}

\begin{subtable}{\columnwidth}
\centering\small
\resizebox{\columnwidth}{!}{
\begin{tabular}{c||c|c|c}
\bf Id & \bf Tweet & \bf \texttt{PL} & \bf \texttt{TL} \\ 
\hline\hline
3564 & @USER @USER @USER Do not engage with idiots, they'll bring you down to their level and beat you with experience.
& \texttt{OFF} & \texttt{NOT} \\
\hline
3725 & This heartburn is disgusting. & \texttt{OFF} & \texttt{NOT}
\end{tabular}}
\caption{Misclassified not offensive examples.}
\label{stab:misnot-example}
\end{subtable}

\caption{Misclassified examples. \texttt{PL}: predicted label, \texttt{TL}: true label}
\label{tab:misclassified-examples}
\end{table}



\subsection{Conclusion}
\label{sec:conclusion}

This paper explores the performance of six individual transformer-based models and their ensemble model for the task of offensive language identification in social media.
Default BERT-Base and RoBERTa-Base individual fine-tuning models are adapted to establish the strong baselines for the ensemble model. 
Sentence representations from six individual models are concatenated and fed into the linear decoder to make binary decision for the ensemble model. 
Our ensemble model with higher dropout shows significant improvements on accuracy, up to 8.6\%, on the dev set than baseline models. 
However, it performs worse than the baseline model \texttt{B-FT} and original ensemble model \texttt{E} on the test set, which has a 92.153\% accuracy. 
It may be caused by model overfitting and data imbalance, which are the problems we need to take into consideration in future experiments.

\section*{Acknowledgments}

We gratefully acknowledge the support of the AWS Machine Learning Research Awards (MLRA).
Any contents in this material are those of the authors and do not necessarily reflect the views of AWS.

\bibliographystyle{coling}
\bibliography{coling2020}

\end{document}